\pdfoutput=1
\documentclass[11pt]{article}
\usepackage[final]{acl}
\usepackage{todonotes}
\usepackage{graphicx}
\usepackage{amsmath}
\usepackage{times}
\usepackage{latexsym}
\usepackage{booktabs}
\usepackage{enumitem}
\usepackage[T1]{fontenc}
\usepackage[utf8]{inputenc}
\usepackage{csquotes}

\usepackage{microtype}
\usepackage{inconsolata}
\usepackage{multirow}
\usepackage{amssymb}
\usepackage{makecell}
\usepackage{lscape}
\usepackage{adjustbox}
\usepackage{url}
\usepackage{subcaption}
\usepackage{placeins}

\usepackage{cleveref}
\crefname{section}{\S}{\S}
\crefname{table}{Table}{Tables}
\crefname{figure}{Fig.}{Figs.}
\crefname{algorithm}{Alg.}{}
\crefname{ALC@unique}{Line}{Lines}
\crefname{equation}{Eq.}{Eqs.}
\crefname{appendix}{App.}{Apps.}
\crefformat{section}{\S#2#1#3} 

\newcommand{\Zero}{Zero}
\newcommand{\Few}{Few}
\newcommand{\CoT}{CoT}
\newcommand{\Text}{Text}
\newcommand{\Graph}{Graphs}

\newcommand{\TAG}{TAG}
\newcommand{\Full}{Full}
\newcommand{\Small}{Small}

\title{TAG\textendash{}EQA: Text\textendash{}And\textendash{}Graph for Event Question Answering via Structured Prompting Strategies}
\author{
Maithili Kadam \quad Francis Ferraro \\
University of Maryland, Baltimore County \\
\texttt{mkadam1@umbc.edu, \quad ferraro@umbc.edu}
}

\begin{document}
\maketitle
\begin{abstract}
Large language models (LLMs) excel at general language tasks but often struggle with event-based questions—especially those requiring causal or temporal reasoning. We introduce \textbf{TAG-EQA} (\underline{T}ext-\underline{A}nd-\underline{G}raph for \underline{E}vent \underline{Q}uestion \underline{A}nswering), a prompting framework that injects causal event graphs into LLM inputs by converting structured relations into natural-language statements. TAG-EQA spans nine prompting configurations, combining three strategies (zero-shot, few-shot, chain-of-thought) with three input modalities (text-only, graph-only, text+graph), enabling a systematic analysis of when and how structured knowledge aids inference. On the TORQUESTRA benchmark, TAG-EQA improves accuracy by ~5\% on average over text-only baselines, with gains up to ~12\% in zero-shot settings and ~18\% when graph-augmented CoT prompting is effective. While performance varies by model and configuration, our findings show that causal graphs can enhance event reasoning in LLMs without fine-tuning, offering a flexible way to encode structure in prompt-based QA.\footnote{Code and data available at \url{https://github.com/MaithiliKadam4/TAG-EQA}}
\end{abstract}

\section{Introduction}
\begin{figure}[!t]
    \centering
    \includegraphics[width=0.95\linewidth]{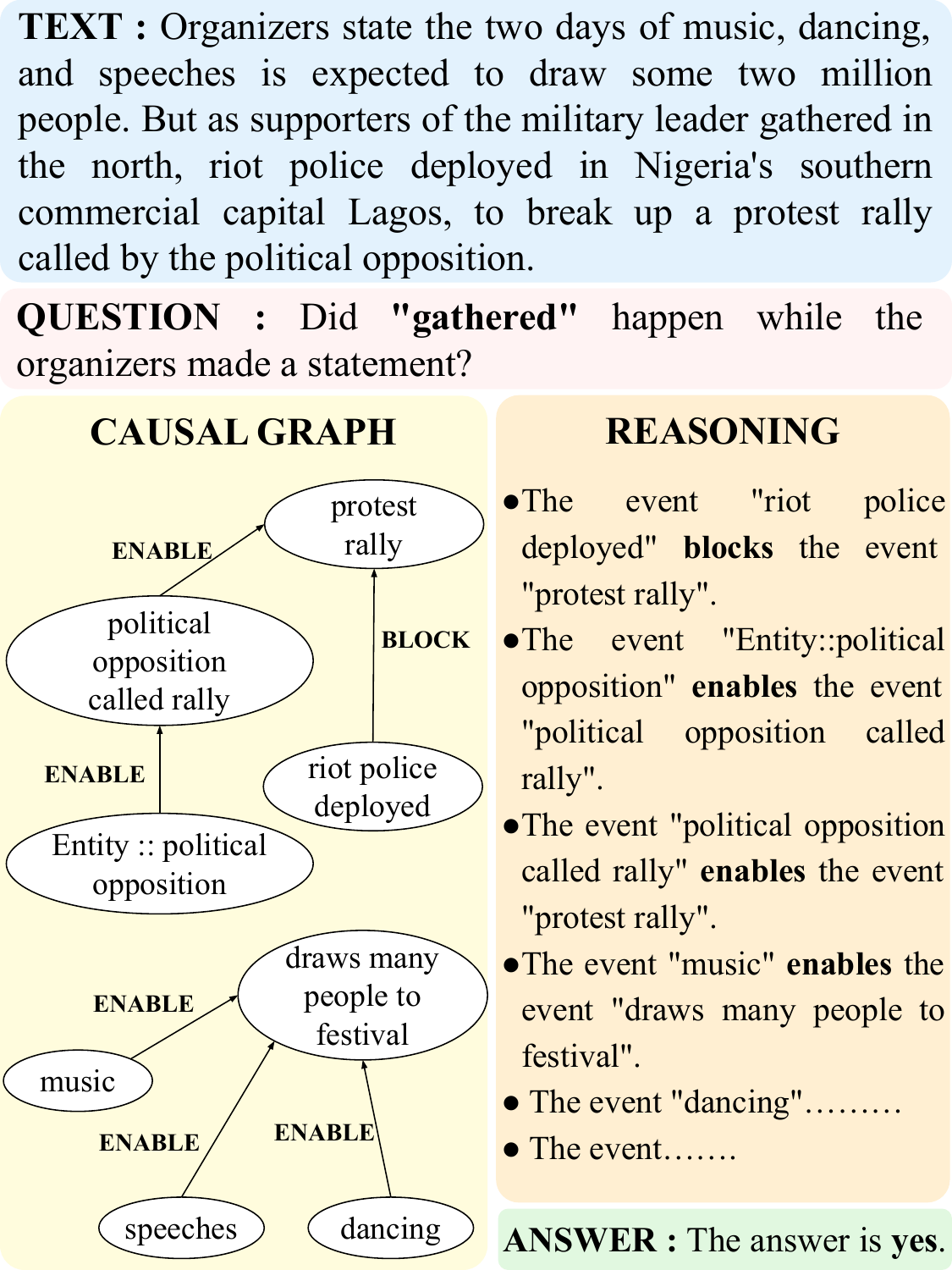}
    \caption{
        Illustrative example from the TORQUESTRA dataset. 
        \textbf{Top:} Narrative passage and a binary event--based question. 
        \textbf{Left:} Annotated causal graph showing \textsc{Enable} and \textsc{Block} relations between events. 
        \textbf{Right:} A step--by--step reasoning trace that follows the graph to support causal inference. Together, the graph and reasoning highlight how structured event relations enable models to answer questions that require indirect causal chaining.
    }
    \label{fig:example}
\end{figure}

Consider the text in Figure~\ref{fig:example}: \emph{“Organizers state the two days of music, dancing, and speeches is expected to draw two million people. But as supporters gathered... riot police deployed...”}. When asked, \emph{``Did the protesters \textup{\textsc{gather}} while the organizers \textup{\textsc{made a statement}}?''}
, answering correctly requires chaining events: $\mathit{music} \rightarrow \mathit{draw\_crowd} \rightarrow \mathit{gather}$, while recognizing that $\mathit{riot\_police\_deployed} \dashv \mathit{organizers\_state}$, where $\rightarrow$ denotes an “enables” relation and $\dashv$ denotes a “blocks” relation. 

Such questions require structured event reasoning, where causal graphs make dependencies explicit by surfacing \textsc{Enable} and \textsc{Block} relations that go beyond surface cues \cite{regan2023causalschema, chambers2008unsupervised, dunietz2020interpretable, jain2023language, chi2024unveiling}. Without structure, LLMs often rely on shallow lexical patterns and miss deeper event logic.

\textbf{We explore how structured causal knowledge can guide large language models in reasoning about events.} Specifically, we introduce \textbf{TAG--EQA}--\textbf{T}ext--\textbf{A}nd--\textbf{G}raph for \textbf{E}vent \textbf{Q}uestion \textbf{A}nswering--a prompting framework that converts causal event graphs into natural language cues and embeds them directly into the prompt. Rather than fine--tuning the model, TAG--EQA steers its inference by aligning causal structure with prompt format, enabling models to reason more coherently about event dynamics. It spans nine prompting configurations, combining three strategies (zero--shot, few--shot, and chain--of--thought) with three input modalities (\emph{text--only}, \emph{graph--only}, and \emph{text+graph}). While this space is broad, our analysis reveals that causal graphs are especially effective when paired with reasoning-oriented prompts such as chain--of--thought. See Section~\ref{sec:method} for full details.

In our experiments on the TORQUESTRA dataset \cite{regan2023causalschema}, TAG--EQA improves accuracy by approximately 5\% over text-only baselines, with gains rising to 12\% in zero-shot and 18\% in chain-of-thought settings. To better understand where structure helps, we group questions into thirteen semantic categories—such as \textit{causal}, \textit{temporal}, and \textit{hypothetical} reasoning—and find that graph-based prompts are particularly effective for \textit{causal chains}, \textit{temporal dependencies}, and \textit{counterfactual what-if scenarios}, where structured event interactions are central to answering correctly. Because these experiments rely on gold human-annotated graphs, the reported numbers should be interpreted as an upper bound on the benefit of structured input; robustness to automatically induced or noisy graphs remains future work.

Our contributions are as follows:
\begin{itemize}[nosep]
  \item We introduce \emph{TAG-EQA}, a prompting framework that incorporates causal event graphs into LLM inputs via natural-language serialization—without requiring model fine-tuning.
  \item We evaluate nine prompting configurations across three strategies and three input types, using T5-XXL, Qwen-32B, and GPT-3.5/4o.\footnote{GPT-3.5 is used for non-reasoning prompts (\Zero{} and \Few{}), while GPT-4o is used for reasoning (\CoT{}) due to its stronger multi-step inference ability.}
  \item We examine how causal graphs and reasoning traces interact, and when they improve model performance.
  \item We report accuracy trends across thirteen semantic question types to identify where structured and/or reasoning-based input helps the most.
\end{itemize}

\section{Related Work}
Prior work on event modeling, causal reasoning, and prompt engineering has independently advanced narrative QA. We synthesize these strands by embedding structured causal graphs into prompt formats to guide event-centric inference in LLMs.

\subsection{Event Modeling}
Narrative understanding has long relied on modeling event relations such as causality, enablement, and sequence. Early work induced event chains using verb–argument frames \citep{chambers2008unsupervised}, while later approaches inferred causal links from raw text without explicit structure \citep{dunietz2020interpretable}. TORQUESTRA \citep{regan2023causalschema} builds on this by aligning QA pairs with human-annotated causal graphs, enabling evaluation of structured reasoning in context.

We build on these efforts by treating enable and block relations as first-class prompt components. Each edge is serialized into a natural language sentence, allowing LLMs to ground their reasoning in structured temporal and causal dependencies.

\subsection{Cause-Effect Graphs and Causal Reasoning}
Causal reasoning from text remains a significant challenge for large language models (LLMs), which often conflate correlation with causation \citep{yamin2024failure}. Early methods extracted causal links using pattern-based heuristics \citep{radinsky2012predicting}, while later approaches employed pretrained language models to infer implicit dependencies from raw text \citep{dunietz2020interpretable}. More recent work has shown that explicitly incorporating cause–effect graphs can improve question answering on narrative and commonsense tasks \citep{roy2024causal, bethany2024enhancing}. However, most prior efforts emphasize direct or temporal links, leaving finer-grained structures underutilized.

However, enabling (A enables B) and blocking (C blocks D) relations remain underexplored despite their value in modeling conditional constraints and counterfactuals. We address this by formalizing them into natural-language prompts that explicitly guide LLM reasoning.

\subsection{Prompt Engineering and Chain-of-Thought Reasoning}
Prompt engineering enables pretrained language models to perform new tasks without parameter updates, leveraging \Zero - and \Few - shot in-context learning \citep{petroni2019language, brown2020language}. Chain-of-thought (\CoT) prompting extends this approach by encouraging step-by-step reasoning through natural language traces \citep{wei2022chain}. Enhancements such as self-consistency decoding and automatic CoT generation, aim to improve reliability and reduce dependence on handcrafted examples \citep{wang2023selfconsistency, zhang2023automaticcot}.

Although \CoT{} prompting has shown strong results in arithmetic and symbolic tasks \citep{wei2022chain, kojima2022large}, its use in structured, event-based inference remains limited. We explore this intersection by aligning CoT prompts with causal graphs—letting models reason over explicitly structured event dynamics across prompt formats.

\section{Method} \label{sec:method}
TAG-EQA investigates whether structured causal knowledge and explicit reasoning can improve event-based question answering (QA) when delivered through prompt design. We vary two orthogonal factors:  
(1) the \emph{prompting strategy}—\Zero{}, \Few{}, or \CoT{}, and  
(2) the \emph{input modality}—\Text{}, \Graph{}, or \TAG{} (text and graph combined).  
This results in nine prompting configurations, each combining a reasoning style with one or more input sources. We evaluate these configurations across three instruction-tuned LLMs (T5-XXL, Qwen-32B, and GPT-3.5/4o) to understand how prompt structure and content influence QA accuracy. 

\begin{figure*}[t]
    \centering
    \includegraphics[width=0.95\textwidth, trim=0 15 0 0, clip]{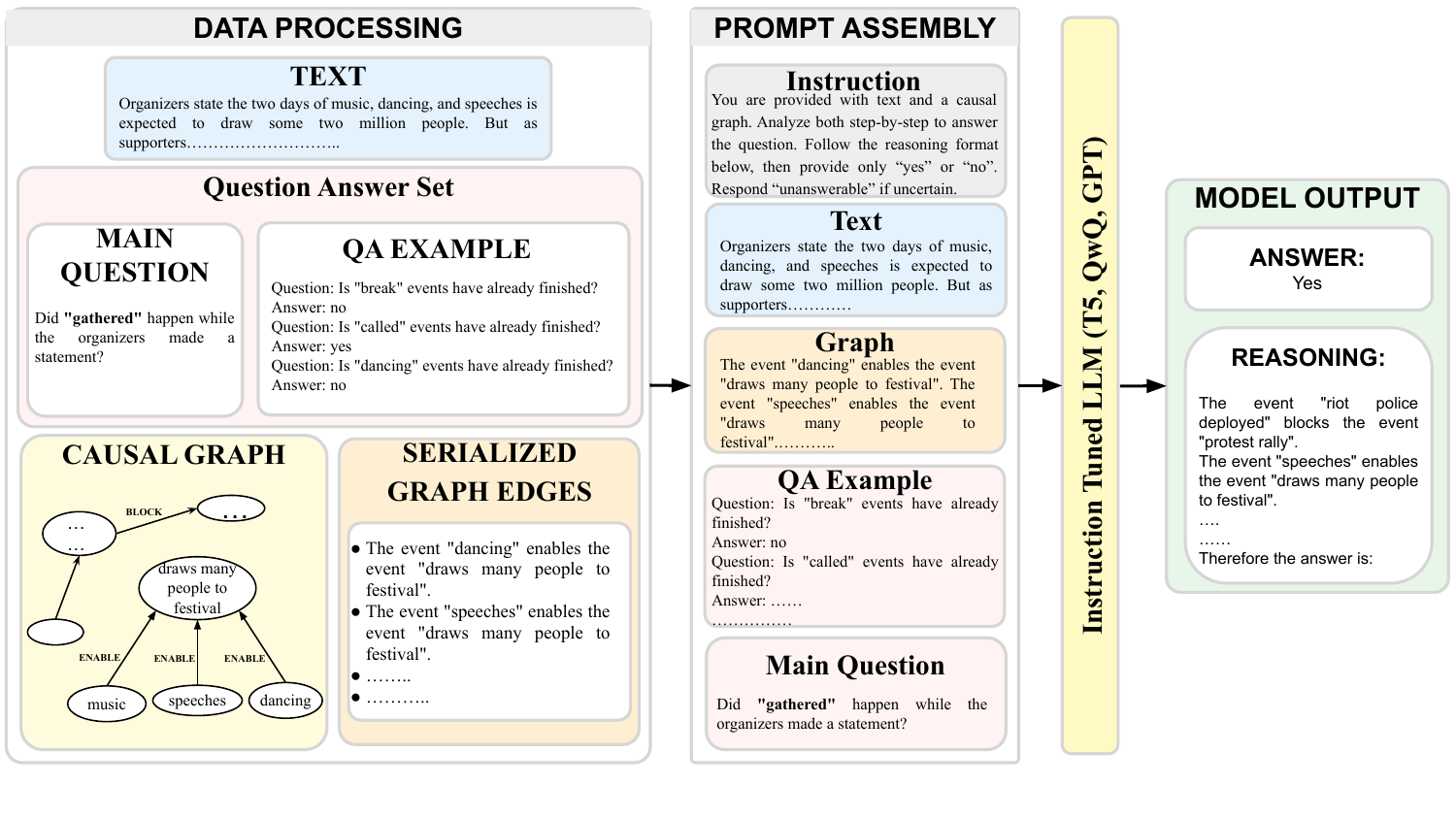}
    \caption{
        \textbf{Overview of our QA prompting pipeline for \TAG{} + \CoT{} configuration.}  
        From left to right: a narrative passage and associated causal graph are processed into a structured input. The causal graph is serialized into natural-language edges (yellow), and the original passage text is retained (blue). Prompt assembly combines task instructions, the text, the graph, in-context QA examples, and the main question into a single input to the instruction-tuned LLM (T5, QwQ, or GPT). The model produces both a yes/no answer and a step-by-step reasoning trace grounded in the causal structure (green).
    }
    \label{fig:methodology}
\end{figure*}

Figure~\ref{fig:methodology} provides a visual overview of our prompting pipeline. Prompts are constructed by combining a narrative passage, a natural-language representation of a causal graph (if present), and optionally, demonstration QA examples or intermediate reasoning traces. See Section~\ref{sec:prompt-track-configs} for full details.

\subsection{Task Formulation}
Each instance consists of a short passage $P$, a yes/no question $Q$ about events in $P$, and optionally a causal event graph $G$--either an \emph{instance} or \emph{schema} graph--encoding directed \textsc{ENABLES}/\textsc{BLOCKS} dependencies. 
An \textsc{ENABLES} edge ($A \rightarrow B$) indicates that event $A$ provides a prerequisite or supportive condition for event $B$ to occur, 
while a \textsc{BLOCKS} edge ($C \dashv D$) denotes that event $C$ prevents, interrupts, or otherwise inhibits event $D$. 
The model must output ``yes'' or ``no.'' In \CoT{} prompts, it must first produce a natural-language reasoning trace, then the final answer.

\subsection{Dataset: TORQUESTRA}
We use the TORQUESTRA dataset \citep{regan2023causalschema} to construct prompts for event-based QA grounded in causal and temporal structure. Each instance provides a short narrative passage, a yes/no question, and one or more directed causal graphs with \texttt{ENABLES}/\texttt{BLOCKS} edges. We generate prompts for all nine configurations by combining QA pairs with the corresponding passage and/or a verbalized version of the graph (i.e., each edge serialized into a natural-language sentence such as “Event A enables Event B”), formatted according to the selected prompting strategy (\Zero{}, \Few{}, or \CoT{}) and input modality (\Text{}, \Graph{}, or \TAG{}).

All prompts are derived from the human-refined subset (TORQUESTRA\textsubscript{human}), which provides gold-standard causal graphs. Figure~\ref{fig:example} illustrates a typical example: the passage, graph, and question are used to build the prompt, although the figure content is for exposition only and not used verbatim.

Our filtered \Full{} split contains 477{,}549 QA instances, balanced across strategies and input types. To support ablations and cost-sensitive models, we also define a \Small{} subset of 1{,}024 instances, stratified by question category and prompting configuration. Unless otherwise noted, results are reported on the \Full{} set, with \Small{} results shown separately for GPT-based models.

Prompt length varies considerably by configuration. For example, \Zero{}--\Text{} prompts average 95.2 tokens, while \CoT{}--\TAG{} prompts reach 336.8 tokens on average, with reasoning traces contributing ~30.7 tokens. These differences affect both model performance and context length constraints.

As shown in Table~\ref{tab:promptstats}, 26.5\% of answers are “yes” and 73.5\% are “no.” Each causal graph omits approximately 5.3 events on average, requiring inference over missing links—a key motivation for evaluating the utility of structured prompts.

\begin{table}[t!]
    \centering
    \scriptsize
    \setlength{\tabcolsep}{4pt}
    \begin{adjustbox}{max width=\columnwidth}
    \begin{tabular}{lcccc}
      \toprule
      \textbf{Track Name} & \textbf{Strategy} & \textbf{Modality} & \textbf{Avg.\ Prompt Length} & \textbf{Reason Length} \\
      \midrule
      \Zero{}--\Text{}  & \Zero{} & \Text{}  & 95.2  & --   \\
      \Zero{}--\Graph{} & \Zero{} & \Graph{} & 80.6  & --   \\
      \Zero{}--\TAG{}   & \Zero{} & \TAG{}   & 138.0 & --   \\
      \Few{}--\Text{}   & \Few{}  & \Text{}  & 121.7 & --   \\
      \Few{}--\Graph{}  & \Few{}  & \Graph{} & 178.0 & --   \\
      \Few{}--\TAG{}    & \Few{}  & \TAG{}   & 242.8 & --   \\
      \CoT{}--\Text{}   & \CoT{}  & \Text{}  & 229.2 & 30.7 \\
      \CoT{}--\Graph{}  & \CoT{}  & \Graph{} & 287.6 & 30.7 \\
      \CoT{}--\TAG{}    & \CoT{}  & \TAG{}   & 336.8 & 30.7 \\
      \bottomrule
    \end{tabular}
    \end{adjustbox}
    \caption{\textbf{Prompt lengths for each TAG--EQA track:} Prompt lengths (tokens) across the three strategies (\Zero{}, \Few{}, \CoT{}) and input modalities (\Text{}, \Graph{}, \TAG{}). \CoT{} prompts include explicit reasoning traces.}
    \label{tab:promptstats}
\end{table}

\subsection{Prompt-Track Configurations}
\label{sec:prompt-track-configs}
TAG-EQA combines three prompting strategies with three input modalities, yielding a 3×3 grid of nine prompt configurations (e.g., \Zero{}--\Text{}, \Few{}--\Graph{}, \CoT{}--\TAG{}) evaluated in Section~\ref{sec:results}. Strategies differ in how much supervision or explicit reasoning they include; modalities differ in whether the model receives natural language text, a structured graph, or both.

Each strategy is paired with one input modality:
\begin{description}
[leftmargin=1.2em,
  style=sameline,
  noitemsep,
  topsep=0pt
]
  \item[\Text{}]: the narrative passage only,
  \item[\Graph{}]: a serialized causal graph representing event dependencies,
  \item[\TAG{}]: both the passage and graph, concatenated.
\end{description}

\paragraph{Zero-shot prompting (\Zero{})}
In the \Zero{} track, the model receives task instructions, the input modality (\Text{}, \Graph{}, or \TAG{}), and the target yes/no question—without demonstrations or reasoning traces. This setting tests whether an LLM can reason directly from the input without prior examples. For instance, using only the \Text{} portion of Figure~\ref{fig:example}, the model must decide whether “gathered” occurred while the organizers made a statement.

\paragraph{Few-shot prompting (\Few{})}  
\Few{} prompts add three in-context demonstrations that match the target configuration. Text-only prompts show how to answer using narrative context; graph-based prompts illustrate how causal structure maps to a yes/no label. \TAG{} prompts include both modalities. This setting provides the model with worked examples aligned to the input type.

\paragraph{Chain-of-thought prompting (\CoT{})}  
\CoT{} prompts build on \Few{} by requesting an explicit reasoning trace. Demonstrations include step-by-step rationales showing how answers are derived from temporal or causal chains. When the graph is present, traces may reference edges (e.g., \texttt{BLOCKS}) or event dependencies. This strategy encourages multi-step inference grounded in structured input.

See Appendix~\ref{appendix:promptformats} for formatting templates across all nine configurations.

\subsection{Causal Graph Integration}
Each causal graph $G$ is verbalized into natural language using one sentence per edge—e.g.,~\textit{“Event A \textsc{enables} Event B.”} or \textit{“Event C \textsc{blocks} Event D.”} Sentences are ordered topologically to preserve causal flow and reduce reference distance. Events are described using surface forms from the original passage to ensure clarity and self-containment.

Apart from the presence or absence of the passage, all other aspects of the prompt remain fixed: task instructions, in-context examples (in \Few{}), and reasoning traces (in \CoT{}) follow a shared scaffold across modalities. This design isolates the effect of graph structure while controlling for phrasing, format, and token budget.

Examples of full \Text{}, \Graph{} and \TAG{} prompts for each strategy track appear in Appendix~\ref{appendix:promptformats}.

\subsection{Model Families and Setup}
We evaluate three instruction-tuned large language model (LLM) families across the full 3×3 TAG-EQA prompt matrix:
\begin{itemize}[leftmargin=1.5em,itemsep=-0.15em]
    \item \textbf{T5-XXL} (Google): 11B encoder–decoder model, pretrained with UL2 and fine-tuned on diverse instructions.  
    \item \textbf{Qwen-32B (QwQ)} (Alibaba): 32B multilingual decoder trained with chat and instruction tuning.  
    \item \textbf{GPT-3.5-Turbo} and \textbf{GPT-4o} (OpenAI): proprietary decoder-only models; GPT-3.5 is used for \Zero{} and \Few{}, while GPT-4o is reserved for \CoT{} evaluation on a smaller subset due to cost.
\end{itemize}

All models use greedy decoding (temperature = 0). Inputs are truncated to model-specific context limits (T5: 1k, Qwen: 2k, GPT: 16k), with graph content prioritized over passage if needed. CoT answers are extracted via regex targeting “Therefore, the final answer is: <yes/no>”.

T5 and Qwen are evaluated on both \Full{} and \Small{} subsets; GPT-3.5 runs \Zero{}/\Few{} on \Full{}, and GPT-4o runs \CoT{} on \Small{} due to API constraints.

\section{Evaluation}
We evaluate \textbf{TAG-EQA} using binary classification accuracy: the percentage of questions answered correctly as “yes” or “no.” Each model is tested across all nine configurations—three prompting strategies (\Zero{}, \Few{}, \CoT{}) × three input modalities (\Text{}, \Graph{}, \TAG{}).

For \CoT{} prompts, we extract the final answer using a regex targeting phrases like “Therefore, the final answer is: yes.” If absent, we fall back to the first standalone yes/no token\footnote{Regex: \texttt{\detokenize{[Tt]herefore,.*answer is: (yes|no)}}}. This ensures consistent evaluation across models with variable output formats.

We report results on both the full TORQUESTRA test set (\Full{}, 477K examples) and a 1,024-instance \Small{} subset used for low-resource and cost-sensitive runs (GPT-4o).

To analyze how structure and reasoning affect performance across reasoning types, we group questions into thirteen semantically grounded clusters derived from TORQUESTRA annotations. These extend the original eight-category taxonomy to include finer-grained types such as \textit{positive}, \textit{negative}, \textit{existential}, and \textit{counterfactual}. Accuracy is reported per cluster and per configuration.

See Appendix~\ref{appendix:clustertables} for full cluster definitions and results.

\FloatBarrier
\section{Results} \label{sec:results}
We evaluate how prompting strategy and input modality affect event-based QA performance across three instruction-tuned LLMs: T5-XXL, Qwen-32B (QwQ), and GPT models (GPT-3.5 and GPT-4o). Each model is tested under nine prompting configurations (\Zero{}/\Few{}/\CoT{} × \Text{}/\Graph{}/\TAG{}). T5 and Qwen are evaluated on both the full TORQUESTRA test set (\Full{}) and a 1,024-example subset (\Small{}). GPT-3.5 and GPT-4o are evaluated only on the \Small{} subset: GPT-3.5 for \Zero{} and \Few{} (non-reasoning), and GPT-4o for \CoT{} (reasoning), due to API cost and throughput constraints.

Across models, \Few{}-shot prompting consistently outperforms \Zero{}-shot in \Text{}-only settings. \CoT{} prompting yields mixed results: QwQ achieves the highest overall accuracy (74.8\%) with \TAG{}-\CoT{}, while T5 performs best with \Few{}-\Text{}. For T5, accuracy drops when \CoT{} is combined with structured input, suggesting difficulty integrating reasoning traces and graph content.

\Graph{} inputs significantly enhance zero-shot and CoT performance for QwQ, sometimes outperforming \TAG{} inputs. However, modality fusion does not always help: \TAG{} configurations often underperform compared to single-modality prompts, particularly for T5. GPT results remain relatively flat across input types, with \Zero{}-\Text{} (58.7\%) performing best for GPT-3.5, and modest gains from CoT in GPT-4o.

These findings highlight the importance of model-aware prompt design: performance gains depend not just on adding structure or reasoning, but on whether a given model can effectively integrate them.

\subsection{Does reasoning (\CoT{}) improve performance over \Zero{} or \Few{}-shot using just text?} \label{sec:results-prompt-type}

We begin by comparing \Zero{}, \Few{}, and \CoT{} prompting under \Text{}--only inputs. As shown in Table~\ref{tab:textonly_prompttype}, \Few{} consistently outperforms \Zero{} across models and data sizes. For example, T5 improves from 54.1\% to 58.5\% on \Full{}, and QwQ improves from 66.8\% to 70.2\%.

\begin{table}[t!]
\centering
\small
\setlength{\tabcolsep}{6pt}
\begin{tabular}{llccc}
\toprule
\textbf{Model} & \textbf{Dataset} & \textbf{Zero} & \textbf{Few} & \textbf{CoT} \\
\midrule
\multirow{2}{*}{\textbf{T5}} 
 & Full  & 54.08 & 58.49 & 55.21 \\
 & Small & 52.64 & 59.47 & 55.96 \\
\midrule
\multirow{2}{*}{\textbf{QwQ}}
 & Full  & 66.78 & 70.21 & 65.77 \\
 & Small & 68.03 & 78.32 & 73.70 \\
\midrule
\multirow{2}{*}{\textbf{GPT}}
 & Full  &   -     &    -     &   -     \\
 & Small & 58.65 & 52.73 & 72.28   \\
\bottomrule
\end{tabular}
\caption{
\textbf{Prompt-Type Accuracy (\%) Comparison on \Text--Only Input.}
Each model is evaluated on the \Full{} and \Small{} TORQUESTRA subsets. \Few{}--shot prompting consistently outperforms \Zero{}--shot on both scales. \CoT{} shows limited gains on \Full{}, but outperforms \Few{} on \Small{} for QwQ and GPT-4o. GPT results are based on \Small{} only due to cost constraints.
}
\label{tab:textonly_prompttype}
\end{table}

\CoT{} prompting shows mixed effects in the absence of graph input. On the \Full{} set, it underperforms \Few{} for both T5 and QwQ. However, on the \Small{} subset, \CoT{} provides noticeable gains: QwQ improves from 70.2\% to 73.7\%, and GPT-4o achieves 72.3\%, outperforming GPT-3.5’s \Few{} score of 52.7\%.

These results suggest that chain-of-thought reasoning can help in low-data settings or with models tuned for step-by-step reasoning, such as GPT-4o. Still, \Few{} remains the most reliable strategy when using plain text alone—especially on larger test sets. GPT results are limited to the \Small{} subset: \Zero{} and \Few{} use GPT-3.5, while \CoT{} uses GPT-4o.

\subsection{Are \Graph{} helpful when used alone or combined with \Text{}?}
\label{sec:results-input-modality}

We evaluate the effect of input modality—\Text{}, \Graph{}, and \TAG{}—under both \Zero{} and \Few{} prompting. 

As shown in Table~\ref{tab:modality_results}a, \Graph{}--only inputs consistently outperform \Text{}--only across models. 
For instance, QwQ improves from 66.8\% (\Text{}) to 78.8\% (\Graph{}), and T5 gains from 54.1\% to 58.0\%. 
Combining \Text{} and \Graph{} in a \TAG{} prompt further improves performance for QwQ (74.5\%) but reduces accuracy for T5 (52.6\%). 
On the \Small{} subset, GPT shows limited variation across modalities -- ranging from 56.8\% to 58.8\% -- indicating relative insensitivity to structured input in zero-shot settings. 
Overall, these results suggest that causal graphs substantially aid zero-shot inference, but modality fusion (Text+Graph) can introduce conflicts depending on the model.

Table~\ref{tab:modality_results}b shows that \Few{}--shot prompting generally boosts absolute performance compared to \Zero{}. 
For example, QwQ achieves its highest score (79.4\%) with \TAG{}, confirming that demonstrations and graph input are complementary. 
GPT gains from Graph input (62.9\%) compared to Text-only (52.7\%), while T5 shows limited or negative gains from structure, dropping from 59.5\% (Text) to 50.8\% (TAG). 
These results suggest that few-shot demonstrations amplify the utility of structured graphs for models like QwQ, and GPT, but highlight integration challenges for T5.

\begin{table*}[t!]
\centering
\small
\setlength{\tabcolsep}{6pt}
\begin{subtable}{0.48\linewidth}
\centering
\begin{tabular}{llccc}
\toprule
\textbf{Model} & \textbf{Dataset} & \textbf{\Text{}} & \textbf{\Graph{}} & \textbf{\TAG{}} \\
\midrule
\multirow{2}{*}{\textbf{T5}} 
 & Full  & 54.08 & 57.96 & 52.58 \\
 & Small & 52.64 & 58.89 & 52.50 \\
\midrule
\multirow{2}{*}{\textbf{QwQ}}
 & Full  & 66.78 & 78.77 & 74.48 \\
 & Small & 68.03 & 68.09 & 67.77 \\
\midrule
\multirow{2}{*}{\textbf{GPT}}
 & Full  &   -     &    -     &   -     \\
 & Small & 58.65 & 56.84 & 58.79   \\
\bottomrule
\end{tabular}
\caption{\Zero--shot prompting.}
\label{tab:zeroshot_modality}
\end{subtable}
\hfill
\begin{subtable}{0.48\linewidth}
\centering
\begin{tabular}{llccc}
\toprule
\textbf{Model} & \textbf{Dataset} & \textbf{\Text{}} & \textbf{\Graph{}} & \textbf{\TAG{}} \\
\midrule
\multirow{2}{*}{\textbf{T5}} 
 & Full  & 58.49 & 57.54 & 51.87 \\
 & Small & 59.47 & 57.32 & 50.76 \\
\midrule
\multirow{2}{*}{\textbf{QwQ}}
 & Full  & 70.21 & 70.48 & 79.37 \\
 & Small & 78.32 & 70.51 & 78.10 \\
\midrule
\multirow{2}{*}{\textbf{GPT}}
 & Full  &   -     &    -     &   -     \\
 & Small & 52.73 & 62.99 & 59.28 \\
\bottomrule
\end{tabular}
\caption{\Few--shot prompting.}
\label{tab:fewshot_modality}
\end{subtable}

\caption{
\textbf{Input Modality Accuracy (\%) Comparison.} 
\textbf{(a)} Zero-shot results: \Graph{}--only inputs outperform \Text{}--only for most models, with QwQ showing the largest gains. 
\textbf{(b)} Few-shot results: Demonstrations improve overall accuracy, and combining graphs with examples (\TAG{}) is especially effective for QwQ, and GPT, while T5 struggles with multimodal integration.
}
\label{tab:modality_results}
\end{table*}


\subsection{When reasoning is explicitly used, does adding a \Graph{} help or hurt?}
\label{sec:results-cot-modality}
We now examine the effect of input modality under \CoT{} prompting. As shown in Table~\ref{tab:cot_modality}, \Graph{}--only inputs improve performance for models capable of leveraging structured representations. QwQ achieves its highest accuracy (74.8\%) with \TAG{}, while also showing strong performance with \Graph{}--only input (72.7\%).

T5 shows modest gains from \Graph{} input: on \Full{}, accuracy rises from 55.2\% (\Text{}) to 56.9\% (\Graph{}), but drops to 50.4\% with \TAG{}, suggesting that reasoning traces may conflict with multimodal inputs for models not tuned for integration. This trend persists on the \Small{} subset.

GPT, evaluated only on \Small{}, shows a slight drop in performance with \TAG{} (70.6\%) compared to \Text{}--only input (72.3\%), while \Graph{}--only input yields comparable performance (71.1\%). This suggests that GPT--4o does not consistently benefit from structured input when combined with reasoning traces in zero-shot settings.

Overall, these results suggest that graph-augmented reasoning is most effective when the model can exploit structure natively--QwQ benefits most—while other models struggle to integrate multiple information sources effectively under CoT prompting.

\begin{table}[t!]
\centering
\small
\setlength{\tabcolsep}{6pt}
\begin{tabular}{llccc}
\toprule
\textbf{Model} & \textbf{Dataset} & \textbf{\Text{}} & \textbf{\Graph{}} & \textbf{\TAG{}} \\
\midrule
\multirow{2}{*}{\textbf{T5}} 
 & Full  & 55.21 & 56.85 & 50.35 \\
 & Small & 55.96 & 56.74 & 49.56 \\
\midrule
\multirow{2}{*}{\textbf{QwQ}}
 & Full  & 65.77 & 72.68 & 74.75 \\
 & Small & 73.70 & 71.55 & 72.05 \\
\midrule
\multirow{2}{*}{\textbf{GPT}}
 & Full  &   -     &    -     &   -     \\
 & Small & 72.28 & 71.07 & 70.61   \\
\bottomrule
\end{tabular}
\caption{
\textbf{Input Modality Accuracy (\%) Comparison in \CoT{} Prompting.}
Each model is evaluated using \CoT{} prompting on the TORQUESTRA dataset. T5 and QwQ show modest to strong gains with \Graph{} inputs. QwQ performs best with combined inputs (\TAG{}), while GPT-4o shows minimal benefit from multimodal prompts. GPT is evaluated only on the \Small{} subset due to API constraints.
}
\label{tab:cot_modality}
\end{table}

\subsection{Which prompting strategy works best for each model?}
\label{sec:results-per-model}
To better understand model-specific behavior, we report each model’s highest-scoring configuration across all nine prompt types (\Zero{}/\Few{}/\CoT{} × \Text{}/\Graph{}/\TAG{}) for both the \Full{} and \Small{} TORQUESTRA subsets (Table~\ref{tab:permodel_best}). Each entry reflects the optimal combination of prompting strategy and input modality at a given data scale.

QwQ achieves the highest overall accuracy (79.4\%) on \Full{} with \Few{}+\TAG{}, showing strong ability to integrate demonstrations and graph input. On \Small{}, it performs best with \Few{}+\Text{}, indicating that graph augmentation is less beneficial under data constraints.

T5 reaches its top accuracy with \Few{}+\Text{} on both subsets (58.5\% and 59.5\%), showing a clear preference for demonstrations alone. Performance declines when graph input or reasoning traces are included, consistent with earlier observations.

GPT, evaluated only on \Small{}, performs best with \Zero{}+\Text{} (72.3\%), suggesting that neither examples nor reasoning traces help much in this setup.

Overall, effective prompting varies by model and scale: structure and reasoning help only when the model can integrate them meaningfully.

\begin{table}[t]
\centering
\small
\setlength{\tabcolsep}{5pt}
\begin{tabular}{lllc}
\toprule
\textbf{Model} & \textbf{Dataset} & \textbf{Best Config} & \textbf{Accuracy\%} \\
\midrule
\multirow{2}{*}{\textbf{T5}}    
    & \Full{}  & \Few{} + \Text{}  &  58.49 \\
    & \Small{} & \Few{} + \Text{}  &  59.47 \\
\midrule
\multirow{2}{*}{\textbf{QwQ}}    
    & \Full{}  & \Few{} + \TAG{}   &  79.37 \\
    & \Small{} & \Few{} + \Text{}  &  78.32 \\
\midrule
\multirow{2}{*}{\textbf{GPT}}    
    & \Full{}  &         -         &    -    \\
    & \Small{} & \Zero{} + \Text{} &  72.28 \\
\bottomrule
\end{tabular}
\caption{
\textbf{Best-Prompting Configuration per Model.}
Top-performing strategy and input modality for each model on the \Full{} and \Small{} TORQUESTRA subsets. GPT results are based on the \Small{} set only due to API cost constraints.
}
\label{tab:permodel_best}
\end{table}

\subsection{Do certain question types benefit more from reasoning or graphs?} \label{sec:results-question-type}

We evaluate model accuracy across thirteen question types derived from TORQUESTRA annotations, extending the original eight clusters (see Appendix~\ref{app:cluster-defs} for details). Figure~\ref{fig:cluster_tag} shows accuracy under \TAG{} input—combined text and graph—across Zero-shot, Few-shot, and Graph-CoT prompting. \textbf{QwQ and GPT perform best in causal and temporal categories} such as \textit{causal}, \textit{past}, \textit{positive}, and \textit{temporal\_conflict}, particularly with CoT prompting. Structured input and reasoning traces appear to help these models handle abstract event relationships.

\textbf{QwQ and GPT perform best on structured categories}—such as \textit{causal}, \textit{past}, \textit{positive}, and \textit{temporal\_conflict}—especially when using CoT prompting. Structured input and step-by-step reasoning appear to help these models capture abstract event relationships.

\textbf{T5 performs best with Few-shot prompting}, but its performance drops with Graph-CoT on speculative or underspecified types like \textit{possible}, \textit{present}, and \textit{unknown}, suggesting difficulty integrating structure and reasoning.

\begin{figure}[!t]
    \centering

    \begin{subfigure}{\linewidth}
        \centering
        \includegraphics[width=\linewidth]{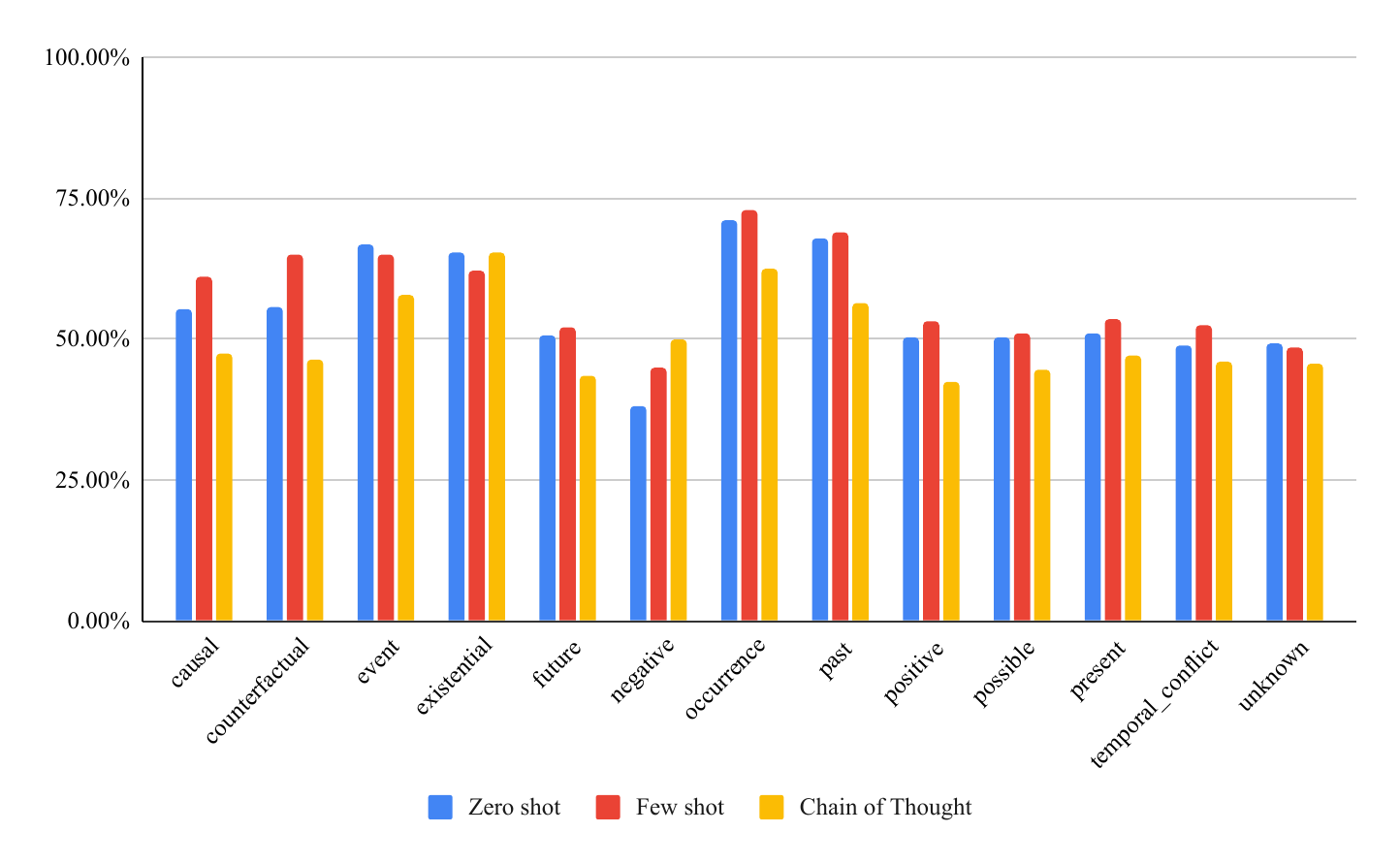}
        \caption{T5 under \TAG{}: \emph{Few\textemdash Text} is strongest overall; \emph{Graph\textemdash CoT} tends to underperform on speculative or underspecified types (\textit{possible}, \textit{present}, \textit{unknown}).}
        \label{fig:cluster_tag_t5}
    \end{subfigure}
  
    \begin{subfigure}{\linewidth}
        \centering
        \includegraphics[width=\linewidth]{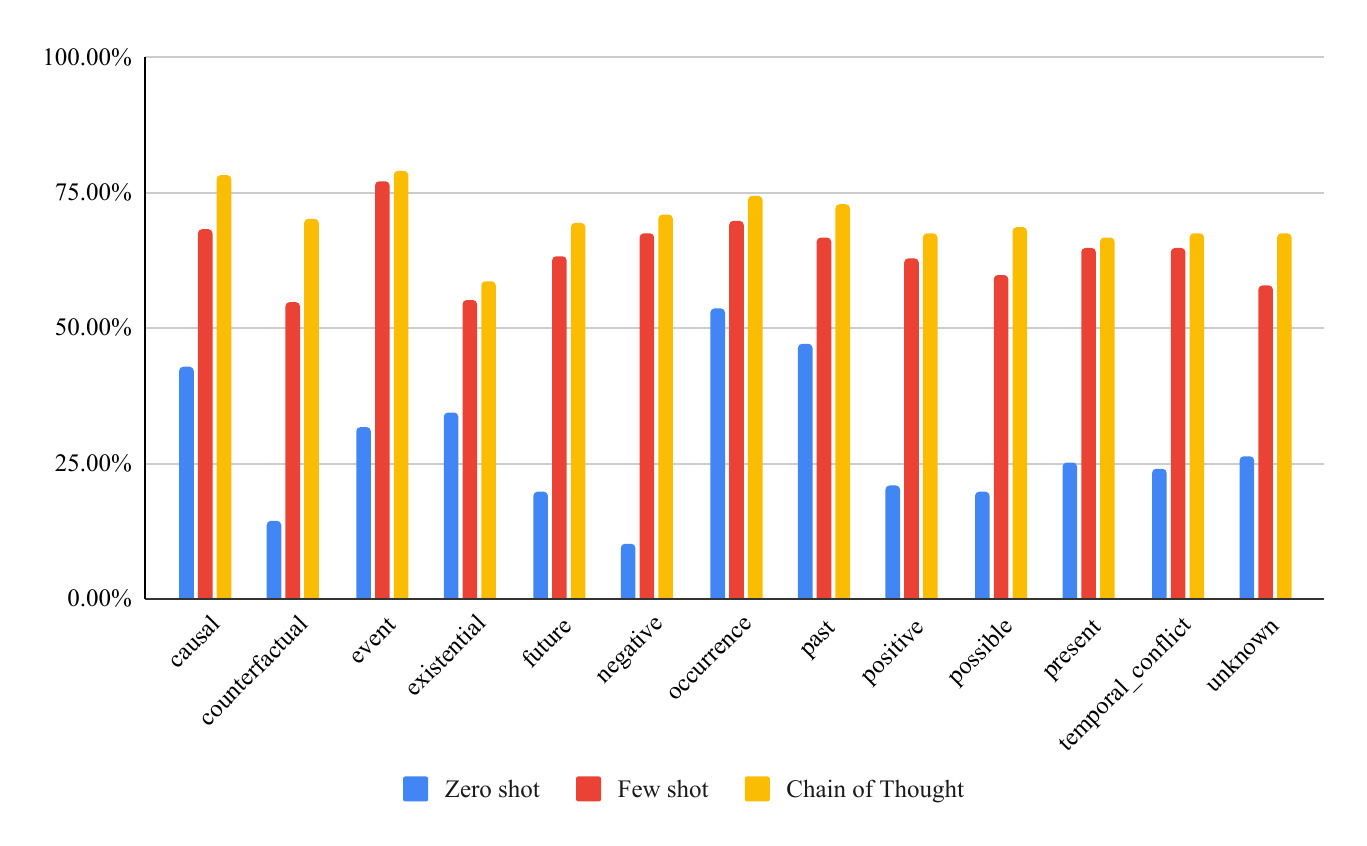}
        \caption{QwQ under \TAG{}: \emph{Graph\textemdash CoT} generally leads on structured types (\textit{causal}, \textit{past}, \textit{temporal\_conflict}); \emph{Few\textemdash Text} remains competitive elsewhere.}
        \label{fig:cluster_tag_qwq}
    \end{subfigure}

    \begin{subfigure}{\linewidth}
        \centering
        \includegraphics[width=\linewidth]{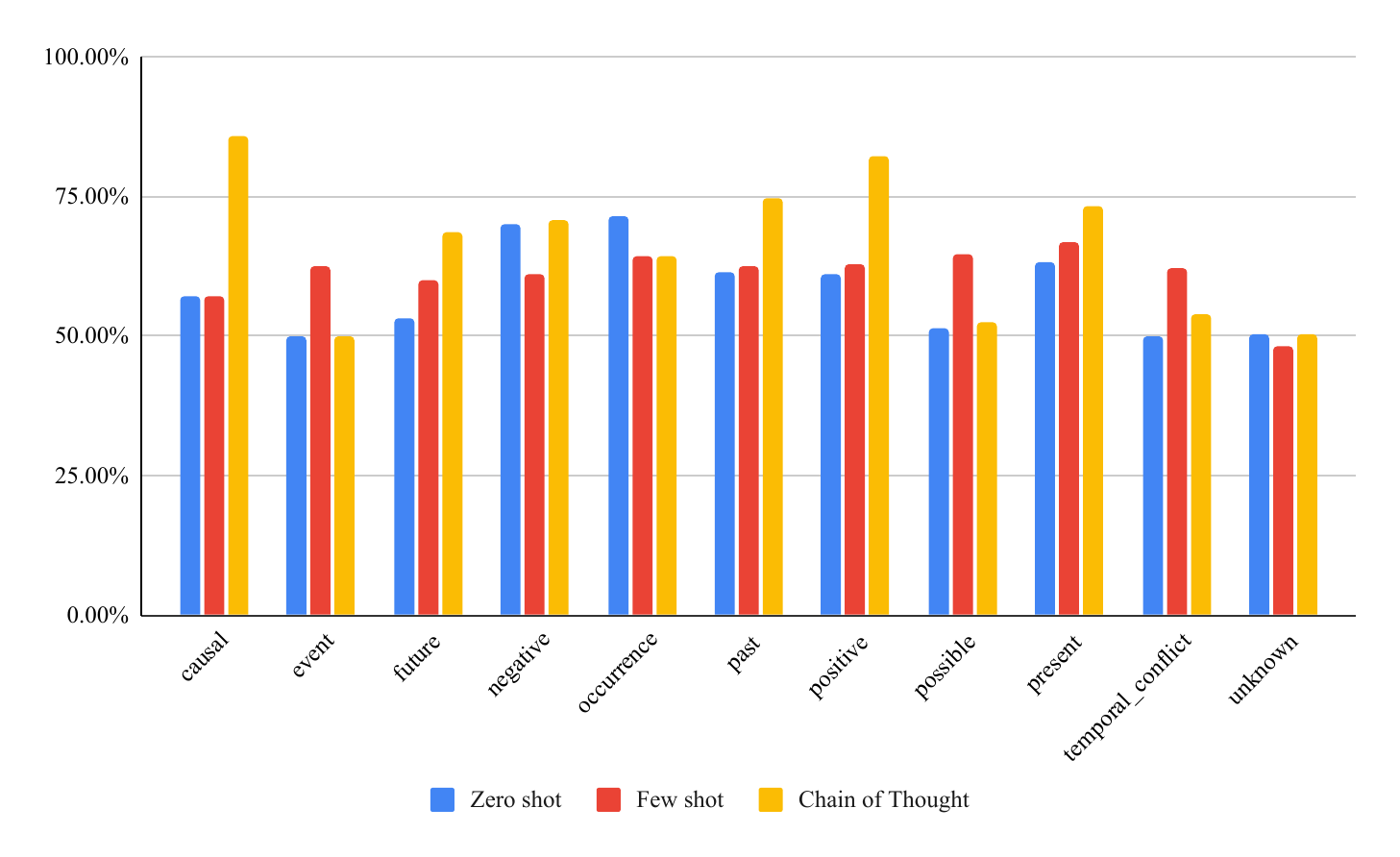}
        \caption{GPT under \TAG{}: \emph{Graph\textemdash CoT} improves relational categories (e.g., \textit{causal}, \textit{temporal\_conflict}); strategy gaps narrow on underspecified types (\textit{possible}, \textit{unknown}).}
        \label{fig:cluster_tag_gpt}
    \end{subfigure}
    
    \caption{
        \textbf{Cluster-wise accuracy under the \TAG{} configuration.}
        Bars denote \emph{Zero\textemdash Text} (blue), \emph{Few\textemdash Text} (red), and \emph{CoT} with \TAG{} (yellow) across thirteen question types. Subfigures (a--c) report T5, QwQ, and GPT respectively. Text--only and Graph--only cluster results appear in Appendix Figures~\ref{fig:cluster_text} and~\ref{fig:cluster_graph}.}
    \label{fig:cluster_tag}
\end{figure}

Appendix Figures~\ref{fig:cluster_text} and~\ref{fig:cluster_graph} show that \Text{} prompts gain from Few-shot examples but struggle with relational types, while \Graph{} prompts provide stronger performance for QwQ in categories like \textit{causal}, \textit{past}, and \textit{event}.

Overall, structured prompting benefits causal and temporal reasoning, with QwQ and GPT showing the strongest gains from graph-augmented CoT. Ambiguous or speculative questions remain difficult across models.

\section{Conclusion}
We introduced TAG-EQA, a systematic framework for evaluating event-based question answering (QA) in large language models (LLMs) using structured causal graphs and reasoning-driven prompting. Our experiments covered nine prompting configurations—three strategies (\Zero{}, \Few{}, \CoT{}) crossed with three input modalities (\Text{}, \Graph{}, \TAG{})—evaluated on three instruction-tuned LLMs: T5, Qwen (QwQ), and GPT models.

Causal graphs consistently improved accuracy on event-centric questions, particularly for relational categories such as \textit{causal}, \textit{past}, and \textit{temporal\_conflict}. QwQ achieved the highest overall performance when combining structure and reasoning (\TAG{}+\CoT{}), while T5 performed best with \Few{}+\Text{} and showed limited gains from structured input. GPT models, evaluated only on a smaller subset, showed moderate benefits from \CoT{} prompting but little sensitivity to input modality. Ambiguous or underspecified categories—such as \textit{possible} and \textit{unknown}—remained challenging across models and prompting styles. These findings highlight both the strengths and limitations of using structured causal input to guide reasoning in LLMs.

\section{Limitations and Future Work}
Our evaluation relies on expert-annotated causal graphs from TORQUESTRA, which provide clean structure but do not reflect the sparsity or noise of automatically induced graphs. The reported numbers should therefore be interpreted as an \emph{upper bound} on the benefit of structured input. Prompt construction is also manually designed—including example selection and reasoning trace format—which may limit generalization to new domains without automation. Performance further varies across models: QwQ benefits most from graph-augmented \CoT{} prompting, whereas T5 and GPT show more modest or inconsistent gains. Due to API cost constraints, GPT models are evaluated only on the \Small{} subset—a full-scale \CoT{} run with GPT-4o would exceed \$950.\footnote{See Appendix~\ref{sec:api-cost} for detailed cost calculations.} Lastly, our binary QA task simplifies causal reasoning and does not capture the complexity of multi-hop inference or generative outputs.

Beyond these constraints, our study is limited to three instruction-tuned LLM families (T5, QwQ, GPT); other architectures may respond differently to structured prompts. We also restrict evaluation to TORQUESTRA, leaving extensions to broader narrative QA datasets (e.g., NarrativeQA, MCTest) for future work. Finally, while we report average prompt lengths, a systematic study of context budget and scaling effects remains open.

Future directions include automated graph construction, robustness to noisy or incomplete graphs, and adaptive graph selection to filter only edges relevant to a query. Extending TAG-EQA with dynamic reasoning traces, instruction tuning for graph-structured \CoT{} prompting, and applications to generative or interactive tasks—such as story simulation, causal forecasting, or decision support—offers promising next steps for leveraging structured knowledge in real-world applications.

\section*{Acknowledgments}
We thank the reviewers for their detailed comments and suggestions. %
Some experiments were conducted on the UMBC HPCF, supported by the National Science Foundation under Grant No. CNS-1920079. %
This material is also based on research that is in part supported by the Army Research Laboratory, Grant No. W911NF2120076, and by DARPA for the SciFy program under agreement number HR00112520301. %
The U.S. Government is authorized to reproduce and distribute reprints for Governmental purposes notwithstanding any copyright notation thereon. The views and conclusions contained herein are those of the authors and should not be interpreted as necessarily representing the official policies or endorsements, either express or implied, of DARPA or the U.S. Government.

\bibliography{latex/custom}

\appendix
\section{Appendix}
\label{sec:appendix}
This appendix provides additional implementation details, example prompts, and full evaluation results referenced in the main paper. We include:
(1) prompt format illustrations, (2) input component breakdowns, (3) per-model prompting results, (4) API cost estimates, and (5) expanded cluster definitions and analysis.

\subsection{Prompt Format Examples}
\label{appendix:promptformats}
We show two full prompt examples in the \CoT{} setting: one using only a causal graph (\Graph{}) and one using both the passage and graph (\TAG{}). These correspond to the instance in Figure~\ref{fig:example}.

\subsubsection{\Graph{} -- CoT Prompt}
\begin{quote}
\small
\texttt{\#\#\# Instruction \#\#\#} \\
\texttt{You are provided with a causal graph and examples showing how to answer. Use only the graph and answer \textquotedblleft yes\textquotedblright{} or \textquotedblleft no\textquotedblright{} only.}

\vspace{0.4em}
\texttt{\#\#\# Graph \#\#\#} \\
\texttt{The event "riot police deployed" blocks the event "protest rally".} \\
\texttt{The event "political opposition" enables the event "political opposition called rally".} \\
\texttt{The event "political opposition called rally" enables the event "protest rally".} \\
\texttt{The event "music" enables the event "draws many people to festival".} \\
\texttt{The event "dancing" enables the event "draws many people to festival".} \\
\texttt{The event "speeches" enables the event "draws many people to festival".}

\vspace{0.4em}
\texttt{\#\#\# Examples \#\#\#} \\
\texttt{Question: Did "protest rally" happen after "riot police deployed"?} \\
\texttt{Answer: no} \\
\texttt{Question: Did "music" cause "draws many people"?} \\
\texttt{Answer: yes}

\vspace{0.4em}
\texttt{\#\#\# Question \#\#\#} \\
\texttt{Did "gathered" happen while the organizers made a statement?} \\
\texttt{\#\#\# Answer \#\#\#}
\end{quote}

\subsubsection{\TAG{} -- CoT Prompt}
\begin{quote}
\small
\texttt{\#\#\# Instruction \#\#\#} \\
\texttt{You are provided with text, a causal graph, and examples showing how to answer. Integrate both and answer \textquotedblleft yes\textquotedblright{} or \textquotedblleft no\textquotedblright{} only.}

\vspace{0.4em}
\texttt{\#\#\# Text \#\#\#} \\
\texttt{Organizers state the two days of music, dancing, and speeches is expected to draw some two million people. But as supporters gathered in the north, riot police deployed in Lagos to break up a protest rally called by the political opposition.}

\vspace{0.4em}
\texttt{\#\#\# Graph \#\#\#} \\
\texttt{[Same graph as above]}

\vspace{0.4em}
\texttt{\#\#\# Examples \#\#\#} \\
\texttt{[Same examples as above]}

\vspace{0.4em}
\texttt{\#\#\# Question \#\#\#} \\
\texttt{Did "gathered" happen while the organizers made a statement?} \\
\texttt{\#\#\# Answer \#\#\#}
\end{quote}

\subsection{Prompt Component Matrix}
\label{appendix:promptcomponents}
Table~\ref{tab:promptmatrix} summarizes which components appear in each of the nine prompting configurations used in TAG-EQA.

\begin{table}[t!]
\centering
\small
\setlength{\tabcolsep}{6pt}
\begin{tabular}{llccc}
\toprule
\textbf{Modality} & \textbf{Input} & \textbf{\Zero} & \textbf{\Few} & \textbf{\CoT} \\
\midrule
\multirow{3}{*}{\Text}
  & Text     & \checkmark & \checkmark & \checkmark \\
  & Graph    &            &            &            \\
  & Examples &            & \checkmark & \checkmark \\
\midrule
\multirow{3}{*}{\Graph}
  & Text     &            &            &            \\
  & Graph    & \checkmark & \checkmark & \checkmark \\
  & Examples &            & \checkmark & \checkmark \\
\midrule
\multirow{3}{*}{\TAG}
  & Text     & \checkmark & \checkmark & \checkmark \\
  & Graph    & \checkmark & \checkmark & \checkmark \\
  & Examples &            & \checkmark & \checkmark \\
\bottomrule
\end{tabular}
\caption{Components used in each prompt configuration.}
\label{tab:promptmatrix}
\end{table}

\subsection{API Cost Estimate}
\label{sec:api-cost}
We compute cost estimates for GPT-3.5 and GPT-4o using OpenAI's May 2025 pricing. Table~\ref{tab:api-cost} shows that a full \CoT{} evaluation with GPT-4o would exceed \$950, so we restrict GPT results to the \Small{} subset.

\begin{table}
\centering
\resizebox{\columnwidth}{!}{
\begin{tabular}{llcc}
\toprule
\textbf{Model} & \textbf{Prompt Type} & \textbf{Total Tokens} & \textbf{Cost (USD)} \\
\midrule
GPT-3.5 & Zero-shot  & 9.7M in / 53k out  & \$4.95 \\
GPT-3.5 & Few-shot   & 12.6M in / 53k out & \$6.36 \\
GPT-4o  & CoT        & 21.1M in / 212M out & \$957.16 \\
\bottomrule
\end{tabular}
}
\caption{Estimated cost to run GPT models on \Full{} dataset.}
\label{tab:api-cost}
\end{table}

\subsection{Per-Model Prompting Results}
\label{appendix:permodeltables}
We report accuracy for each model across all $3\times3$ prompting configurations. These tables complement Section~\ref{sec:results-per-model} and clarify which strategies and modalities are most effective for different architectures.

\paragraph{T5.} Performs best with \Few{}+\Text{}, but degrades when structure or reasoning is added.
\begin{minipage}[t]{0.48\textwidth}
\centering
\small
\begin{tabular}{lccc}
\toprule
\textbf{Prompt Type} & \Text{} & \Graph{} & \TAG{} \\
\midrule
\Zero{}     & 54.1 & 58.0 & 52.6 \\
\Few{}      & 58.5 & 57.5 & 51.9 \\
\CoT{}      & 55.2 & 56.9 & 50.4 \\
\bottomrule
\end{tabular}
\captionof{table}{T5 accuracy across all strategies and modalities.}
\end{minipage}

\paragraph{QwQ.} Excels with \TAG{}+\Few{} and \TAG{}+\CoT{}. Gains are consistent across most settings.
\begin{minipage}[t]{0.48\textwidth}
\centering
\small
\begin{tabular}{lccc}
\toprule
\textbf{Prompt Type} & \Text{} & \Graph{} & \TAG{} \\
\midrule
\Zero{}     & 66.8 & 66.8 & 74.5 \\
\Few{}      & 70.2 & 70.5 & 79.4 \\
\CoT{}      & 65.8 & 72.7 & 74.8 \\
\bottomrule
\end{tabular}
\captionof{table}{QwQ accuracy across all strategies and modalities.}
\end{minipage}

\paragraph{GPT.} Best performance under \CoT{} (GPT-4o). GPT-3.5 shows smaller gains and flat modality sensitivity.
\hfill
\begin{minipage}[t]{0.48\textwidth}
\centering
\small
\begin{tabular}{lccc}
\toprule
\textbf{Prompt Type} & \Text{} & \Graph{} & \TAG{} \\
\midrule
\Zero{}     & 58.7 & 56.8 & 58.8 \\
\Few{}      & 52.7 & 63.0 & 59.3 \\
\CoT{}      & 72.3 & 71.1 & 70.6 \\
\bottomrule
\end{tabular}
\captionof{table}{GPT accuracy across all strategies and modalities.}
\vspace{-0.5em}
\begin{flushleft}
\footnotesize \textit{Note:} All GPT results are reported on the \Small{} subset due to API cost constraints.
\end{flushleft}
\end{minipage}

\subsection{Expanded Cluster Definitions}
\label{appendix:clustertables}
We extend TORQUESTRA's original eight cluster categories into thirteen to better capture event-centric reasoning. Table~\ref{tab:cluster_mapping} aligns our expanded taxonomy with the original groups.

\begin{table}[h]
\centering
\small
\begin{tabular}{@{}ll@{}}
\toprule
\textbf{Expanded Category} & \textbf{Original Cluster} \\
\midrule
causal            & causal \\
counterfactual    & causal (extended) \\
event             & event \\
existential       & event (subtype) \\
future            & future \\
negative          & event (negative polarity) \\
occurrence        & event / temporal \\
past              & past \\
positive          & event (positive polarity) \\
possible          & possible \\
present           & present \\
temporal\_conflict & temporal\_conflict \\
unknown           & unknown \\
\bottomrule
\end{tabular}
\caption{Expanded category mapping.}
\label{tab:cluster_mapping}
\end{table}

\subsection{Cluster-Based Accuracy Analysis}
\label{app:cluster-defs}
We present accuracy trends by question category across all prompting configurations.

\textbf{T5:} Best with \Few{}+\Text{} on most clusters. Accuracy drops with \Graph{}+\CoT{}.

\textbf{QwQ:} Excels with \TAG{}+\CoT{}. Leads in most structured and relational categories.

\textbf{GPT (3.5/4o):} \CoT{} (GPT-4o) performs best across categories like \textit{causal} and \textit{past}; GPT-3.5 (\Zero{}/\Few{}) is stable but less sensitive to modality.

\begin{figure*}[t]
    \centering
    \includegraphics[width=0.32\textwidth]{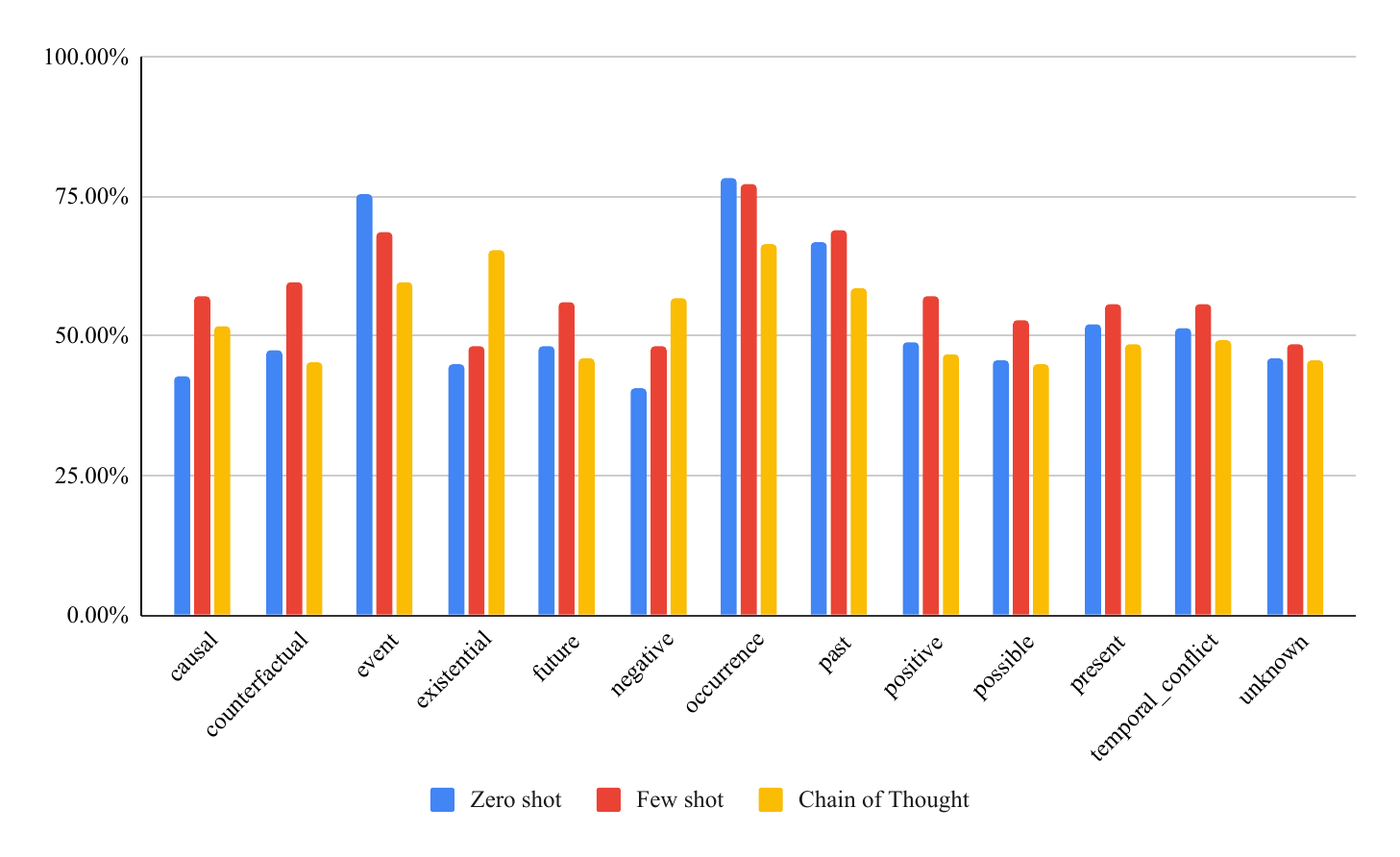}
    \hfill
    \includegraphics[width=0.32\textwidth]{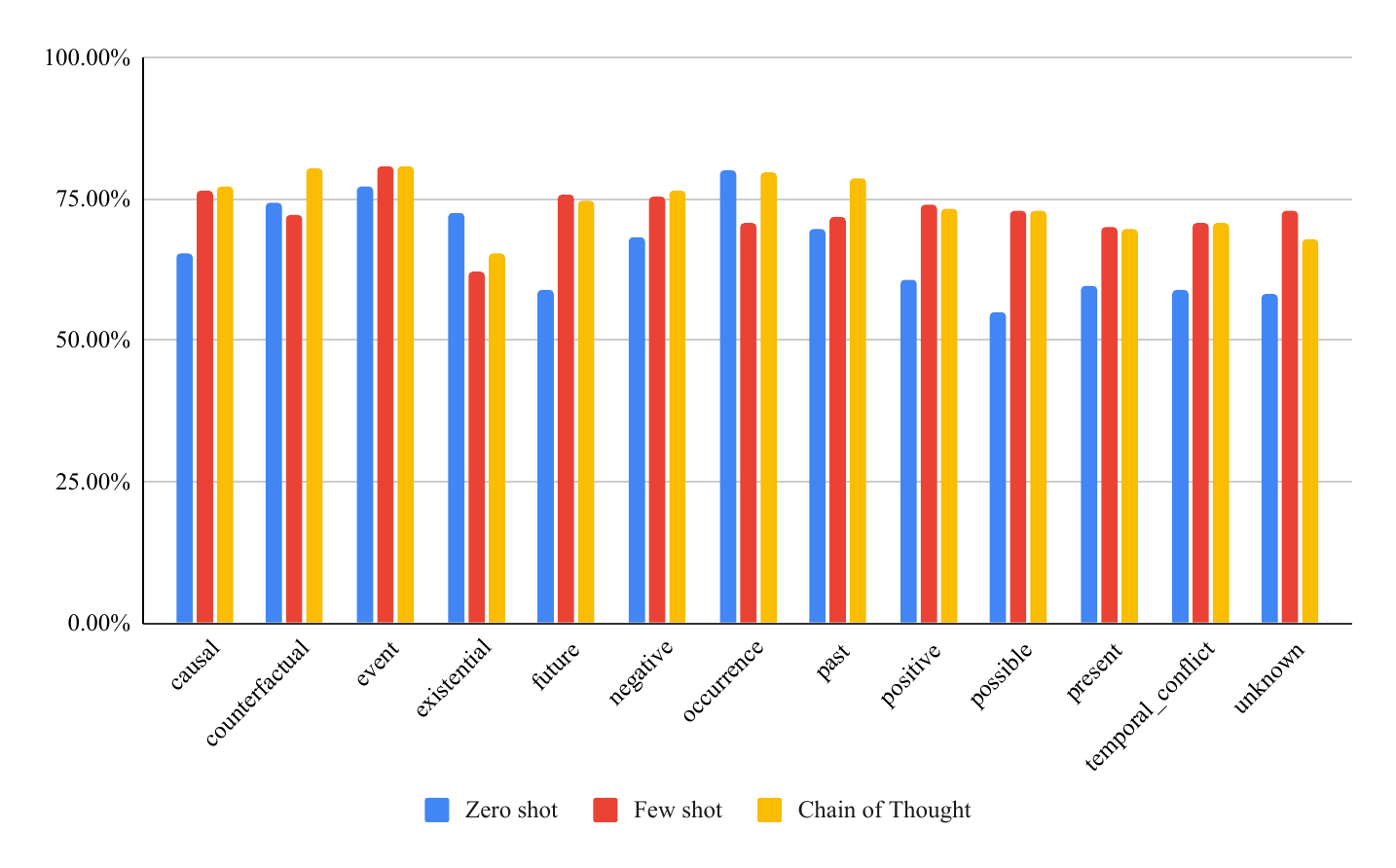}
    \hfill
    \includegraphics[width=0.32\textwidth]{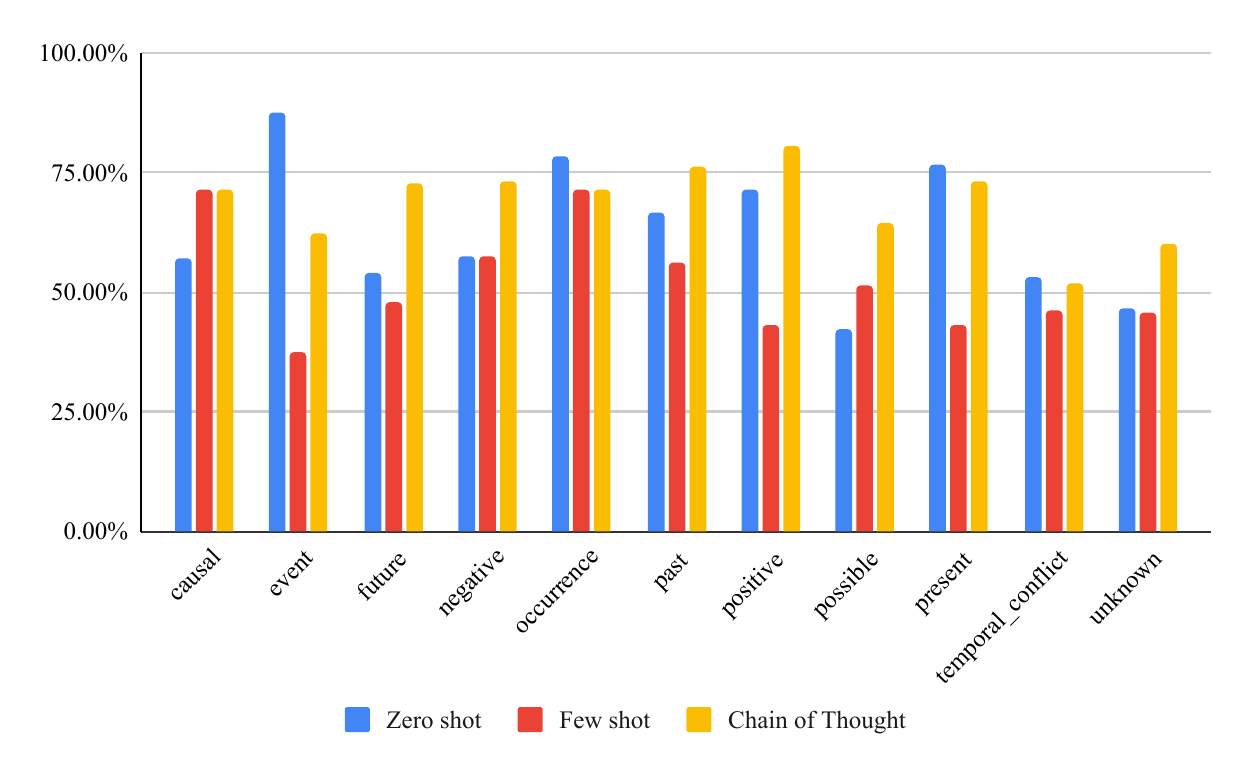}
    \caption{
    \textbf{Cluster-wise Accuracy by Model and Prompting Strategy.}
    Accuracy across thirteen question categories for each model (T5, QwQ, GPT) under three prompting strategies: \Zero{}--\Text{} (blue), \Few{}--\Text{} (red), and \CoT{} with \TAG{} input (yellow). QwQ and GPT benefit most from graph-augmented CoT prompting on structured categories such as \textit{causal}, \textit{past}, and \textit{temporal\_conflict}. T5 performs best with \Few{}--shot but struggles to integrate structure and reasoning. All models show weaker performance on underspecified or speculative categories like \textit{possible} and \textit{unknown}.
    }
    \label{fig:cluster_text}
\end{figure*}

\begin{figure*}[t]
    \centering
    \includegraphics[width=0.32\textwidth]{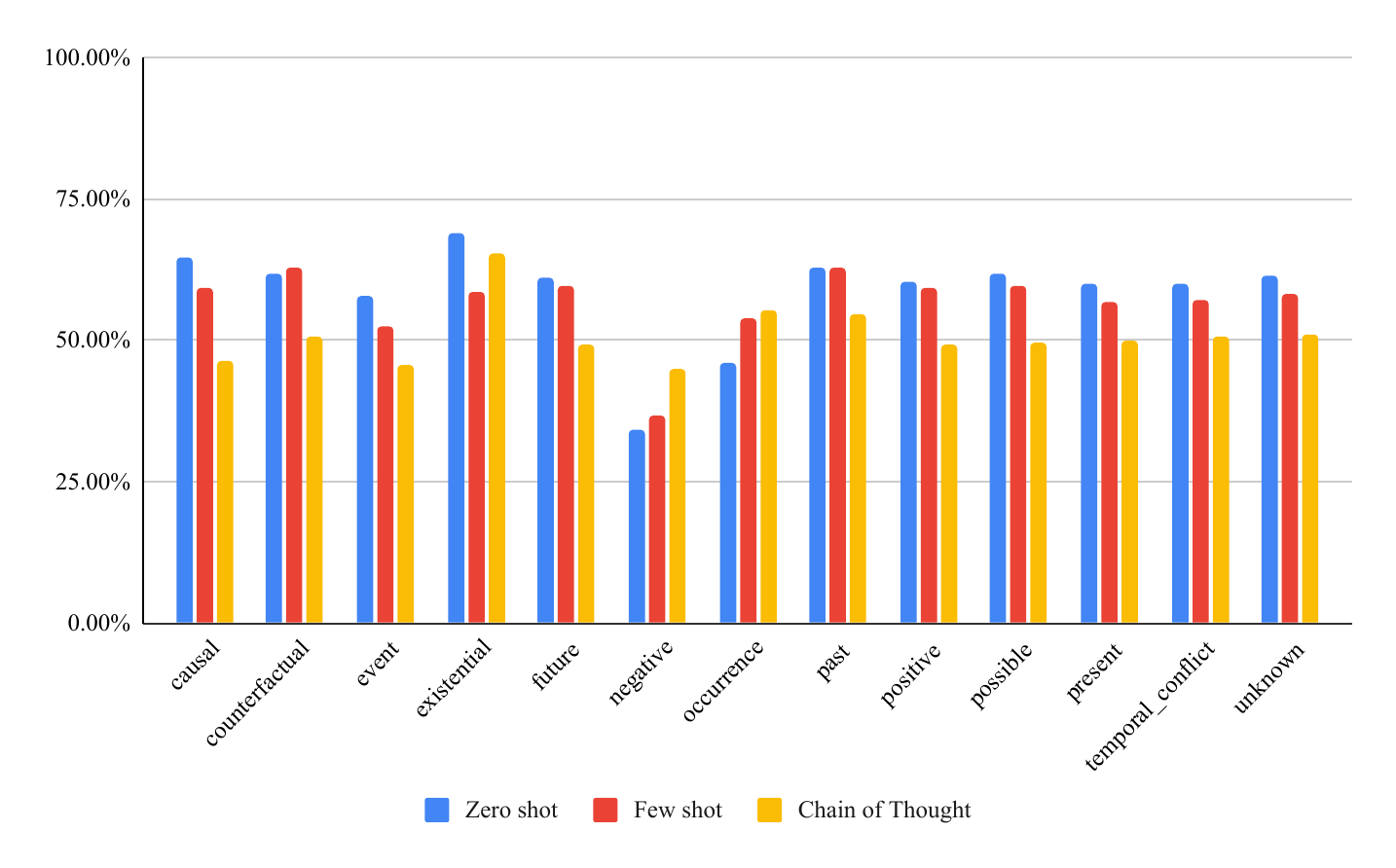}
    \hfill
    \includegraphics[width=0.32\textwidth]{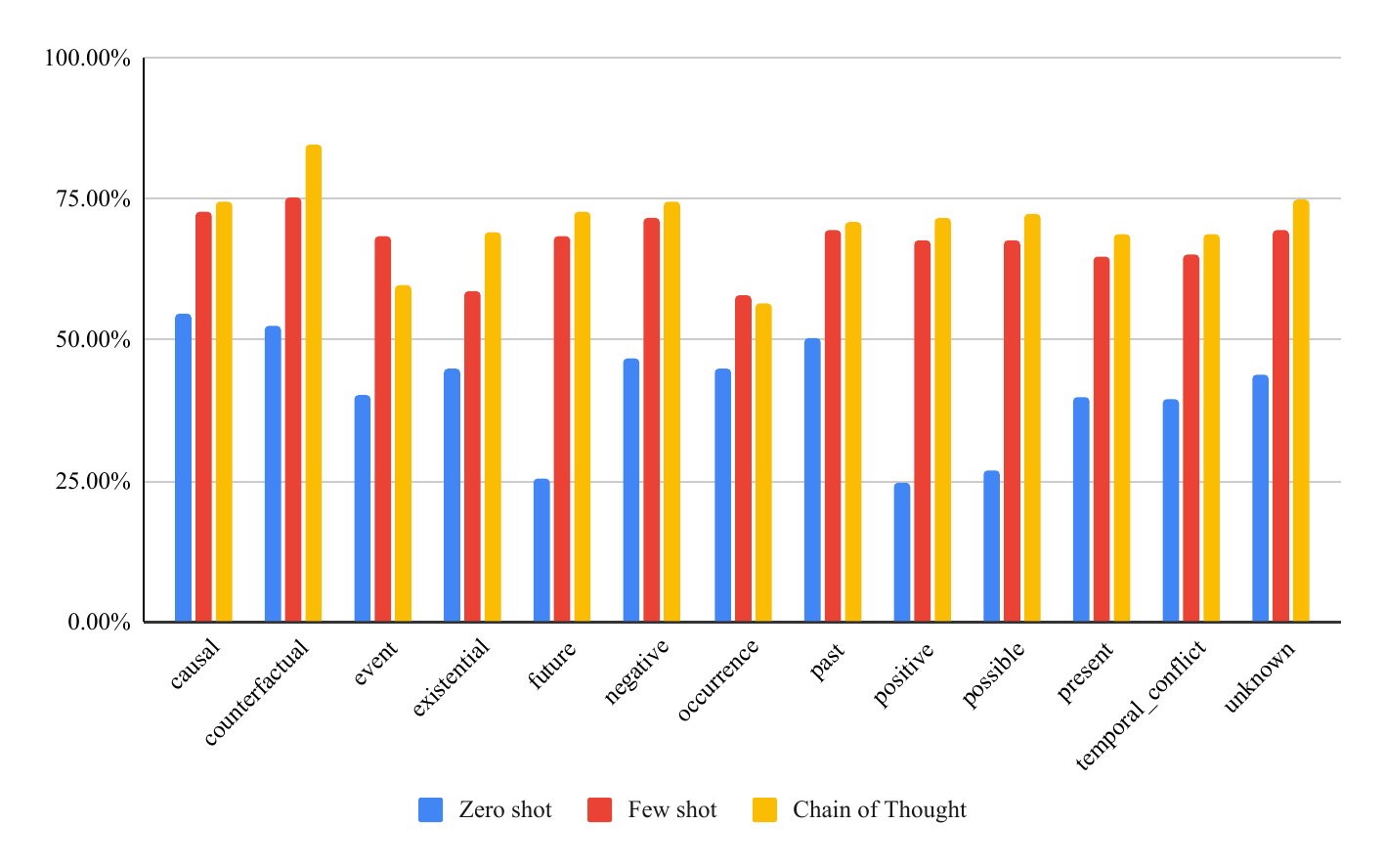}
    \hfill
    \includegraphics[width=0.32\textwidth]{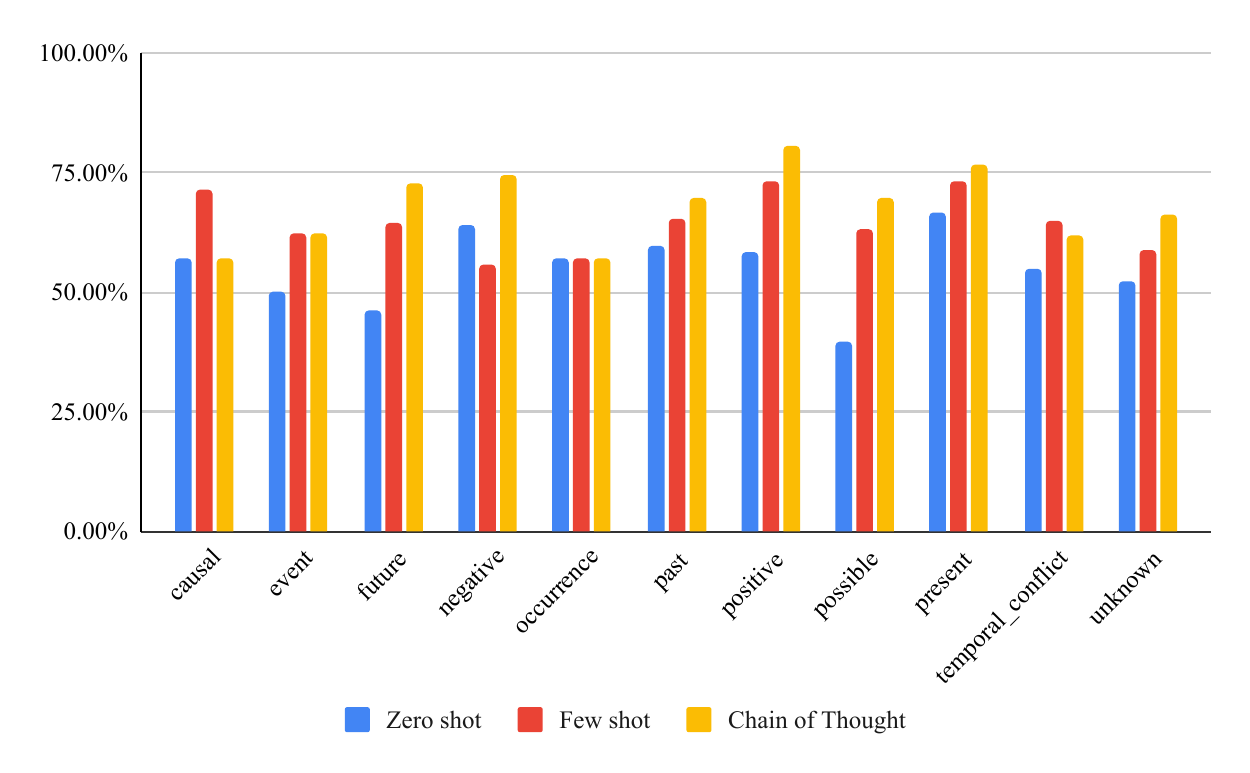}
    \caption{
    \textbf{Cluster-wise Accuracy by Model and Prompting Strategy.}
    Accuracy across thirteen question categories for each model (T5, QwQ, GPT) under three prompting strategies: \Zero{}--\Text{} (blue), \Few{}--\Text{} (red), and \CoT{} with \TAG{} input (yellow). QwQ and GPT benefit most from graph-augmented CoT prompting on structured categories such as \textit{causal}, \textit{past}, and \textit{temporal\_conflict}. T5 performs best with \Few{}--shot but struggles to integrate structure and reasoning. All models show weaker performance on underspecified or speculative categories like \textit{possible} and \textit{unknown}.
    }
    \label{fig:cluster_graph}
\end{figure*}

\end{document}